\documentclass{article}
\usepackage[final]{colm2026_conference}

\usepackage[T1]{fontenc}
\usepackage[utf8]{inputenc}
\usepackage{microtype}
\usepackage{inconsolata}
\usepackage{amssymb}
\usepackage{algorithm}
\usepackage{algpseudocode}
\usepackage{hyperref}
\usepackage{url}
\usepackage{multirow}
\usepackage{wrapfig}
\usepackage{arydshln}
\usepackage{enumitem}
\definecolor{darkblue}{rgb}{0, 0, 0.5}
\hypersetup{colorlinks=true, citecolor=darkblue, linkcolor=darkblue, urlcolor=darkblue}

\usepackage{graphicx}
\usepackage{caption}
\usepackage{booktabs}
\usepackage{array}
\usepackage{adjustbox}
\usepackage{float}
\usepackage{multirow}
\usepackage{colortbl}

\usepackage{amsfonts}
\usepackage{bbm}
\newcommand{\coderepo}[1]{%
  \begingroup
    \hypersetup{urlcolor=red!80!black}%
    \textbf{\texttt{\url{#1}}}%
  \endgroup
}

\usepackage{xcolor}
\definecolor{darkgreen}{RGB}{0,100,0}

\usepackage[most]{tcolorbox}
\tcbset{examplebox/.style={colback=gray!10!white, colframe=blue!50!black, fonttitle=\bfseries}}
\tcbset{tealbox/.style={colback=teal!10!white, colframe=teal!80!black, fonttitle=\bfseries}}
\newtcolorbox[auto counter, number within=section]{examplebox}[2][]{examplebox,#1,title=#2}
\newtcolorbox{example}[1][]{%
  enhanced, breakable,
  colback=blue!5!white, colframe=blue!60!black,
  fonttitle=\bfseries, title=Example, #1
}

\usepackage{cuted}
\usepackage{lineno}

\title{SAKE: Structured Agentic Knowledge Extrapolation for Complex LLM Reasoning via Reinforcement Learning}

\author{
\textbf{Jiashu He\textsuperscript{1}\thanks{Correspondence to \texttt{jiashuhe@seas.upenn.edu}.} \quad Jinxuan Fan\textsuperscript{2} \quad
Bowen Jiang\textsuperscript{1} \quad Ignacio Hounie\textsuperscript{1}} \\
\textbf{Dan Roth\textsuperscript{1} \quad Alejandro Ribeiro\textsuperscript{1}} \\[0.3em]
\textsuperscript{1}University of Pennsylvania \quad
\textsuperscript{2}University of California, Berkeley \\
\texttt{\{jiashuhe, bwjiang, ihounie, danroth, aribeiro\}@seas.upenn.edu} \\
\texttt{jinxuan\_fan@berkeley.edu}
}

\begin{document}

\ifcolmsubmission
\linenumbers
\fi

\maketitle
\lhead{}

\begin{abstract}
Knowledge extrapolation is the process of inferring novel information by combining and extending existing knowledge that is explicitly available. It is essential for solving complex questions in specialized domains where retrieving comprehensive external knowledge is impractical. We propose SAKE (Structured Agentic Knowledge Extrapolation), a RL powered agentic framework that trains LLMs to autonomously retrieve and extrapolate structured knowledge through tool-augmented reinforcement learning. SAKE defines two external KG tools: entity group construction and cross-group triplet retrieval. The model learns to interleave these 2 retrieval tools during a three-turn rollout: extracting key entities, filtering relevant concept groups, and associative reasoning by constructing new triplets through analogy. The entire pipeline is optimized end-to-end with GRPO using a reward that combines output format and answer correctness, teaching the model what to retrieve and how to reason over it. Our experiments proved that SAKE fine-tuned Qwen2.5-7B model surpasses GPT-3.5-Turbo with state-of-the-art agentic KG reasoning on both biomedical (75.4\% vs.\ 70.1\%) and commonsense (81.3\% vs.\ 74.7\%) benchmarks, while reducing token usage by over 90\%. These results demonstrate that associative reasoning over incomplete structured knowledge does \textbf{not} require large models with complex, multi-step prompting, thus can be learned end-to-end by small, open-weight models through reinforcement learning with the right tools and training signal. Our code is available at \coderepo{https://github.com/jxfan99/SAKE}.
\end{abstract}

\section{Introduction}
The ability of inferring novel facts by extending existing knowledge and accessible evidence is an essential aspect of human intelligence, especially when faced with challenging and unfamiliar questions. In this case, humans draw relation between the challenging task and the related known facts, and extrapolate through associative thinking \citep{BEATY2023671, KAUFMAN2009374}. For instance, to answer whether melatonin can help treat insomnia, one may formulate the reasoning process that since melatonin is a hormone and insomnia is a mental disorder, melatonin should be able to treat insomnia as its generally known that mental disorders are treated by hormones. This kind of ability to extend beyond the available knowledge is needed by large language models (LLMs) as well. Despite recent 
advances in structured reasoning and planning 
\citep{xiong2025deliberate, xiong2025enhancing}, deploying AI systems in knowledge-intensive domains \citep{cascella2023evaluating, tian2024opportunities} remains challenging, as neither parametric knowledge from pre-training nor retrieved context at testing time could provide complete chain of reasoning. Relying on either of them leads the model to hallucination and harms the performance.

In this paper, we propose SAKE (Structured Agentic Knowledge Extrapolation), a framework that trains LLMs to perform knowledge extrapolation over structured knowledge graphs (KGs) through tool-augmented reinforcement learning. Concretely, SAKE utilizes the structured nature of the KG as a chain of reasoning that connects the related concepts, and teach the model to \textit{extrapolate} beyond what is explicitly retrievable. Given a query, the model autonomously retrieves relevant KG triplets, identifies structural patterns in them, and constructs new triplets by substituting semantically similar concepts. Therefore, it bridges the gap between incomplete external knowledge and the challenging corner domain question.

SAKE achieves this through an agentic, multi-turn pipeline that the model learns entirely via RL, with no pre-computed retrieval or supervised fine-tuning. The pipeline proceeds as follows. First, the model analyzes the query and extracts key concepts that it needs to analyze. It then invokes an external tool that links these concepts to a KG and returns groups of semantically similar entities. The model filters these groups for relevance, invokes a second tool that retrieves KG triplets connecting the selected groups, and finally performs associative reasoning to construct new triplets by analogy and producing a final answer. All tool invocations are triggered by special tokens that the model generates during its rollout, and the entire trajectory is optimized end-to-end with GRPO \citep{shao2024deepseekmathpushinglimitsmathematical} using a single outcome-based reward, which propagates credits to all intermediate processes including which entities to extract and which groups to retain. This design means that the model discovers, through exploration, that accurate entity extraction, judicious group filtering, and faithful knowledge extrapolation lead to better retrieval and higher accuracy. The full pipeline is tied together through a single outcome-based reward signal.

Prior agentic frameworks have explored integrating retrieval with 
LLM reasoning: Search-R1 \citep{searchR1} and Search-O1 
\citep{searchO1} train models to issue web queries, while 
Graph-R1 \citep{luo2025graphr1} extends this to multi-turn 
hypergraph retrieval. AutoGraph-R1 \citep{tsang2025autographr} 
uses RL to optimize KG construction itself, and 
Knowledgeable-R1 \citep{knowledgeabler1} addresses knowledge 
conflicts between parametric and retrieved knowledge. KG-based 
reasoning methods such as ToG \citep{sun2024thinkongraphdeepresponsiblereasoning} and GraphRAG \citep{edge2025localglobalgraphrag} offer structured retrieval but assume complete KGs, 
while GIVE \citep{he2024give} addresses KG incompleteness through 
inference-time prompting at substantial token cost. SAKE differs 
fundamentally: (1) knowledge extrapolation is \textit{learned} 
through RL rather than engineered through prompts, and (2) 
retrieval and reasoning are \textit{unified} into a single 
end-to-end policy.

Our experiments demonstrate the effectiveness of SAKE across two 
knowledge graphs of vastly different scales: a 135-node UMLS KG 
for biomedical QA (PubMedQA \citep{pubmedqa}, BioASQ \citep{bioasq}, 
ProcessBank \citep{processbank}) and a 844K-node ConceptNet for 
commonsense reasoning (CommonsenseQA). SAKE consistently outperforms 
all baselines across prompting-based reasoning, retrieval-augmented, 
fine-tuning, and agentic framework categories on both model sizes including 3B and 7B. 
Notably, a Qwen2.5-7B model with SAKE surpasses GPT-3.5-Turbo 
paired with a state-of-the-art agentic KG reasoning framework on 
both biomedical (75.4\% vs.\ 70.1\%) and commonsense (81.3\% vs.\ 
74.7\%) benchmarks, while reducing token usage by over 90\% and 
requiring only a single inference pass. Ablation studies further 
confirm that each component of the pipeline: agentic retrieval, 
group filtering, and knowledge extrapolation contributes 
meaningfully, and that the gains arise from structured knowledge 
interaction rather than fine-tuning alone.

\begin{figure*}
    \vspace{-0.5cm}
     \makebox[\textwidth][c]{%
       \includegraphics[width=1.15\textwidth,height=8cm]{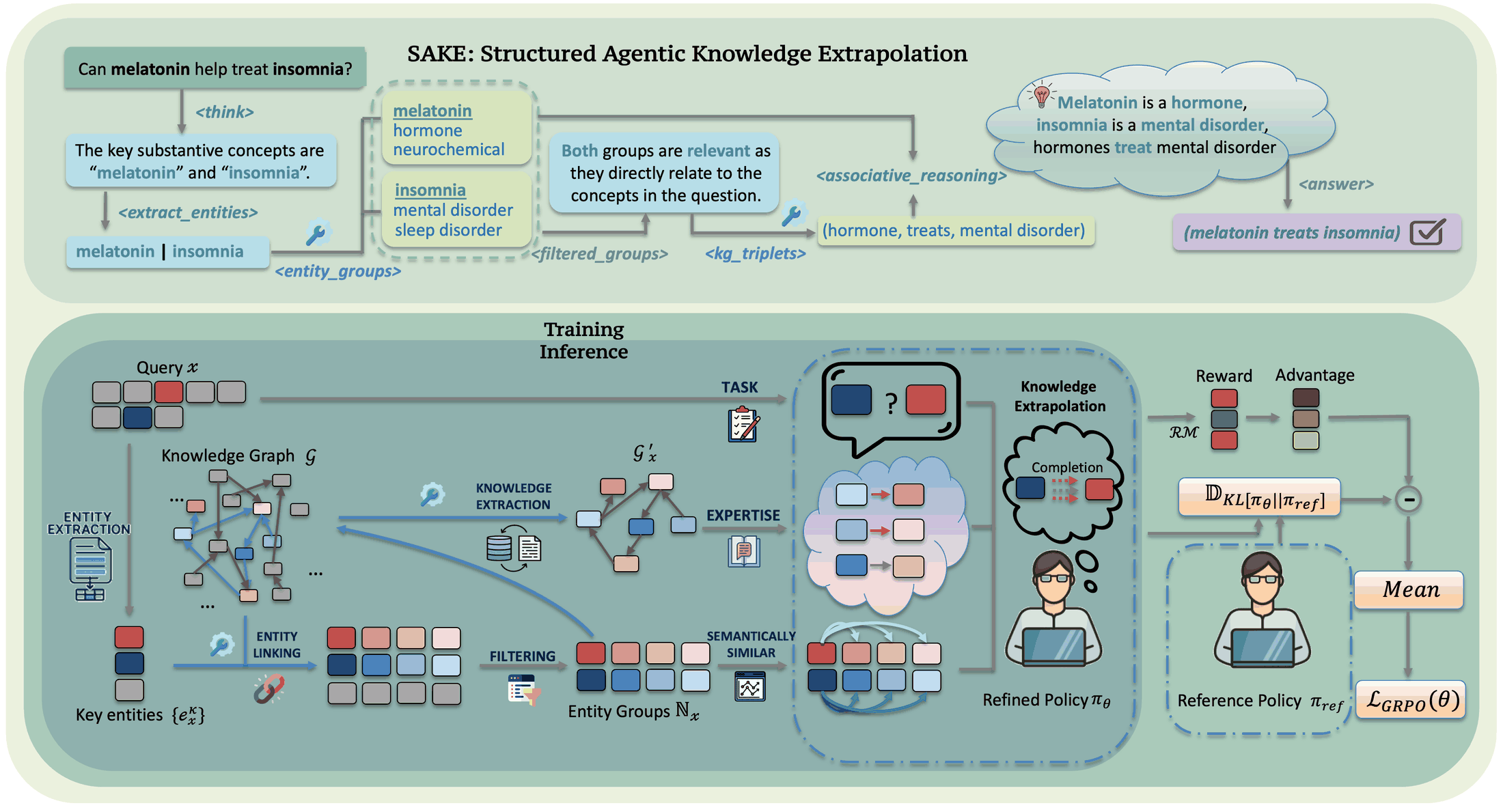}%
     }
    \makebox[\textwidth][c]{%
      \begin{minipage}{1.15\textwidth}
      \captionsetup{font=footnotesize, labelfont=bf}
      \caption{Overview of SAKE. \textbf{Upper:} A concrete example illustrating the three-turn agentic pipeline on a biomedical question. The model generates structured output delimited by special tokens (\texttt{<extract\_entities>}, \texttt{<filtered\_groups>}, \texttt{<associative\_reasoning>}, \texttt{<answer>}), which trigger two external tool calls: entity group construction and cross-group triplet retrieval. The retrieved triplet \textit{(hormone, treats, mental disorder)} is extrapolated into \textit{(melatonin, treats, insomnia)} by substituting semantically similar concepts from the entity groups. \textbf{Lower:} The training and inference framework. For query $x$, the policy $\pi_\theta$ generates a multi-turn rollout by entity extraction, group filtering, and associative reasoning. The rollout generation process is interleaved with tool calls that interact with the knowledge graph $\mathcal{G}$. All tool invocations are triggered by the model's own special-token outputs during the rollout; nothing is pre-computed or provided externally. The entire trajectory is optimized end-to-end with GRPO, where the reward signal flows back through all model decisions including entity extraction, group filtering and knowledge extrapolation.}
      \label{fig::main}
      \end{minipage}%
    }
    \vspace{-0.6cm}
\end{figure*}

\section{Problem Definition}
\label{sec::problem}

\paragraph{Setup.}
Let $\pi_\theta$ be a policy LLM parameterized by $\theta$, $x$ a query with ground-truth answer $a^*$, and $\mathcal{G} = (\mathcal{V}_\mathcal{G}, \mathcal{R}_\mathcal{G}, \mathcal{E}_\mathcal{G})$ a knowledge graph, where $\mathcal{V}_\mathcal{G}$ is the set of entities (nodes), $\mathcal{R}_\mathcal{G}$ is the set of relations, and $\mathcal{E}_\mathcal{G} \subseteq \mathcal{V}_\mathcal{G} \times \mathcal{R}_\mathcal{G} \times \mathcal{V}_\mathcal{G}$ is the set of triplets (edges). We assume the KG may be \textit{incomplete}: there is no guarantee that $\mathcal{E}_\mathcal{G}$ contains triplets that directly answer $x$. The model must therefore \textit{extrapolate} beyond retrieved knowledge to arrive at a correct answer.

\paragraph{Tool-Augmented Environment.}
We define two deterministic tool functions that the policy can invoke during generation. Together, they bridge the gap from the query's conceptual space to the KG's relational space:
\begin{align}
\mathcal{F}_1 &: E_x \;\rightarrow\; \mathbb{N}_x
  & &\text{(entity linking \& group construction)} \label{eq:tool1-def} \\
\mathcal{F}_2 &: \mathbb{N}_x \;\rightarrow\; \mathcal{T} \subseteq \mathcal{E}_\mathcal{G}
  & &\text{(cross-group triplet retrieval)} \label{eq:tool2-def}
\end{align}
where $E_x = \{e_1, \ldots, e_n\}$ denotes the set of key entities extracted from query $x$ (which may not exist verbatim in $\mathcal{V}_\mathcal{G}$), $\mathbb{N}_x = \{G_1, \ldots, G_n\}$ is the collection of entity groups with each $G_i \subseteq \mathcal{V}_\mathcal{G}$ containing $e_i$ and its semantically similar KG concepts, and $\mathcal{T}$ is the set of retrieved cross-group triplets. $\mathcal{F}_1$ maps from the query space to the KG node space: it links each query entity to its nearest neighbors in $\mathcal{V}_\mathcal{G}$, handling the common case where query-specific terminology does not appear in an incomplete KG. $\mathcal{F}_2$ maps from the KG node space to the KG edge space, retrieving all triplets in $\mathcal{E}_\mathcal{G}$ that connect entities across different groups. Both tools are deterministic and external to the policy---they are not parameterized by $\theta$ and produce no gradient signal.

\paragraph{Multi-Turn Trajectory.}
We define a \textit{turn} as a contiguous segment of model-generated 
tokens, bounded by a special token that triggers a tool call 
(or by the end-of-sequence token). A single rollout consists of three 
turns interleaved with two tool calls, producing a trajectory:
\begin{equation}\label{eq:trajectory}
y = \bigl(\,\underbrace{y^{(1)}}_{\text{extract}},\;\underbrace{o^{(1)}}_{\mathcal{F}_1},\;\underbrace{y^{(2)}}_{\text{filter}},\;\underbrace{o^{(2)}}_{\mathcal{F}_2},\;\underbrace{y^{(3)}}_{\text{reason}}\,\bigr)
\end{equation}
where each model turn $y^{(t)}$ is generated autoregressively, conditioned on the query and all preceding turns and tool outputs:
\begin{equation}\label{eq:autoregressive}
y^{(t)} \sim \pi_\theta\!\left(\,\cdot\;\middle|\;x,\;y^{(1)},\,o^{(1)},\;\ldots,\;y^{(t-1)},\,o^{(t-1)}\right)
\end{equation}
Tool outputs are computed deterministically from model-generated content: $o^{(1)} = \mathcal{F}_1\bigl(\mathrm{Parse}(y^{(1)})\bigr)$ and $o^{(2)} = \mathcal{F}_2\bigl(\mathrm{Parse}(y^{(2)})\bigr)$, where $\mathrm{Parse}(\cdot)$ extracts structured content from the model's tagged output. Concretely, $\mathrm{Parse}(y^{(1)})$ yields $E_x$ from the \texttt{<extract\_entities>} tags, and $\mathrm{Parse}(y^{(2)})$ yields a subset $S \subseteq \{1,\ldots,n\}$ of selected group indices from the \texttt{<filtered\_groups>} tags.

\paragraph{Objective.}
We optimize the policy end-to-end over the full trajectory:
\begin{equation}\label{eq:RL}
\max_{\pi_\theta}\;\mathbb{E}_{x \sim \mathcal{D},\,y \sim \pi_\theta(\cdot \mid x)}\bigl[R(y, a^*)\bigr] - \beta\,D_{\mathrm{KL}}\!\bigl(\pi_\theta(\cdot \mid x)\,\|\,\pi_{\mathrm{ref}}(\cdot \mid x)\bigr)
\end{equation}
where $R(y, a^*)$ is the reward function (detailed in 
Section~\ref{sec:reward}), $\pi_{\mathrm{ref}}$ is a fixed reference 
policy, and $\beta$ controls the KL penalty. The policy is conditioned 
only on the query $x$: the retrieved entity groups $\mathbb{N}_x$ and 
triplets $\mathcal{T}$ are not pre-computed or provided as fixed inputs, 
but are determined by the model's own outputs during the rollout. 
Token-level losses are computed only over model-generated tokens 
$(y^{(1)}, y^{(2)}, y^{(3)})$; tool outputs $(o^{(1)}, o^{(2)})$ are 
deterministic and masked from the loss (as described by the binary mask 
$m$ in Algorithm~1). However, because the \textit{content} of tool 
outputs depends on what the model generated in preceding turns, the 
outcome-based reward $R(y, a^*)$ implicitly supervises all upstream 
decisions---the model learns that better entity extraction and group 
filtering lead to more useful retrievals and ultimately higher accuracy.

\section{SAKE}
\label{sec::method}

SAKE equips LLMs with associative reasoning over structured knowledge through an agentic, tool-augmented framework trained end-to-end with reinforcement learning. As illustrated in Figure~\ref{fig::main}, the model interacts with a knowledge graph (KG) through two external tools during each rollout, and learns via RL, that when and how to invoke them effectively to answer complex queries.

The framework proceeds in three generation turns interleaved with two tool calls: the model first extracts key entities from the query (Turn~1), invokes Tool~1 to obtain entity groups, filters the groups for relevance (Turn~2), invokes Tool~2 to retrieve cross-group triplets, and finally performs associative reasoning to produce an answer (Turn~3). Nothing is pre-computed or provided by a human pipeline: the model itself decides which entities to extract, which groups to retain, and how to reason over the retrieved knowledge. The entire multi-turn trajectory is optimized end-to-end with reinforcement learning. We describe each component below.

\subsection{Agentic KG Interaction via Tool Use}
\label{sec::tools}

SAKE defines two deterministic tools that serve as the model's interface to the knowledge graph. These tools are invoked during the rollout: the model generates structured output tags, execution is paused, the tool computes a result, and the result is injected back into the rollout sequence for the model to continue generating.

\paragraph{Tool 1: Entity Group Construction.}
Given a set of extracted entities $E = \{e_1, \ldots, e_n\}$ produced by the model in Turn~1, this tool links each entity to the KG and constructs an entity group containing the entity and its semantically similar siblings.

For each entity $e_i$, let $w$ denote a pre-trained encoder and $\mathcal{V}_\mathcal{G}$ the set of all KG entities. The tool retrieves the $p$ most similar KG concepts:
\begin{equation}\label{eq:group-construction}
Y_i = \underset{\substack{y \in \mathcal{V}_\mathcal{G}}}{\mathrm{Top\text{-}}p}
\;\cos\bigl(w(e_i),\,w(y)\bigr)
\end{equation}
and returns the entity group $G_i = \{e_i\} \cup Y_i$.

The tool returns all groups $\{G_1, \ldots, G_n\}$, wrapped in \texttt{<entity\_groups>} tags and injected into the rollout. Each group provides the model with a neighborhood of semantically related KG concepts, enabling it to bridge the gap between query-specific terminology and the KG's vocabulary.

\paragraph{Tool 2: Cross-Group Triplet Retrieval.}
After the model selects a subset $S$ of relevant groups in Turn~2, this tool retrieves all KG triplets whose head and tail entities belong to different selected groups:
\begin{equation}\label{eq:triplet-retrieval}
\mathcal{T} = \{(h,r,t) \in \mathcal{E}_\mathcal{G} \mid \exists\, (i,j) \in S \times S,\; i \neq j,\; h \in G_i,\; t \in G_j\}
\end{equation}
The retrieved triplets $\mathcal{T}$ are wrapped in \texttt{<kg\_triplets>} tags and injected into the rollout. These cross-group connections provide structured evidence about how concepts from different entity groups relate to each other in the KG, serving as the foundation for associative reasoning.

\subsection{Multi-Turn Rollout}
\label{sec::rollout}

The full rollout procedure is detailed in Algorithm~1 
(Appendix~\ref{appendix:algo}), including the binary token mask $m$ 
that ensures GRPO loss is computed only over model-generated tokens.

\paragraph{Turn 1: Entity Extraction.}
The model receives query $x$ and identifies key biomedical concepts, 
outputting them inside \texttt{<extract\_entities>} tags. The model is 
instructed to extract specific concepts (e.g., diseases, molecules) 
rather than relational words (e.g., ``treatment'', ``cause''), as the 
latter do not correspond to KG entities. Tool~1 then constructs entity 
groups and injects them into the rollout.

\paragraph{Turn 2: Group Filtering.}
The model evaluates each returned entity group's relevance, reasoning 
inside \texttt{<think>} tags and outputting selected group indices 
inside \texttt{<filtered\_groups>} tags. This step is important: 
entity extraction may yield concepts whose KG neighborhoods are noisy, 
and the model learns through RL to discard these before triplet 
retrieval. Tool~2 then retrieves cross-group triplets and injects them.

\paragraph{Turn 3: Associative Reasoning.}
The model reasons over the retrieved triplets and produces a final 
answer. When triplets do not directly answer the query, the model 
constructs new triplets by substituting semantically similar concepts 
from entity groups into existing triplet patterns---e.g., mapping 
\texttt{(hormone, treats, mental\_disorder)} to 
\texttt{(melatonin, treats, insomnia)}. If no triplets are found, the 
model reasons using entity groups and parametric knowledge alone.

\subsection{System Prompt}
\label{sec::prompt}
 
The system prompt (provided in full in Appendix~\ref{appendix:prompt}) defines the three-turn structure and output format for each turn. It instructs the model to extract substantive biomedical concepts (not relational words like ``treatment'' or ``cause''), describes the associative reasoning mechanism of constructing new triplets by substituting semantically similar terms, and provides no task-specific examples. This design is to ensure that reasoning ability emerges from RL training rather than prompt engineering.
\subsection{End-to-End Reinforcement Learning}
\label{sec::RL}

The entire multi-turn trajectory is optimized jointly using 
GRPO~\citep{shao2024deepseekmathpushinglimitsmathematical} within 
the VERL~\citep{verl} framework.

\paragraph{Policy Optimization.}
Token-level losses are computed over all model-generated segments (Turns 1--3), while tool outputs are excluded from the loss computation as they are deterministic. Let $\rho_\theta(y \mid x) = \frac{\pi_\theta(y\mid x)}{\pi_{\theta_{\mathrm{old}}}(y\mid x)}$ denote the importance ratio. The policy is optimized with:
\begin{equation}\label{eq:grpo-objective}
\begin{aligned}
\mathcal{L}_{\mathrm{SAKE}}(\theta)
&= -\mathbb{E}_{x,y}\Bigl[
  \min\bigl(
    \rho_\theta \,\hat{A},\;
    \mathrm{clip}(\rho_\theta,1{-}\epsilon,1{+}\epsilon)\,\hat{A}
  \bigr)\Bigr] \\
&\quad -\,\beta\,\mathbb{E}_{x}\Bigl[
  D_{\mathrm{KL}}\bigl(\pi_\theta \,\|\, \pi_{\mathrm{ref}}\bigr)\Bigr]
\end{aligned}
\end{equation}
where $y = (y^{(1)}, o^{(1)}, y^{(2)}, o^{(2)}, y^{(3)})$ denotes the full multi-turn rollout, $\epsilon$ is the clipping hyperparameter, $\hat{A}$ is the advantage estimated across rollouts within the same group, $\pi_{\mathrm{ref}}$ is a fixed reference policy, and $\beta$ controls the KL penalty.
\subsection{Reward Function}
\label{sec:reward}

We use a weighted additive reward that combines structural compliance with
answer correctness. Given a rollout $y$ produced by the policy $\pi_\theta$
for query $x$ with ground-truth answer $a^*$, the reward $R(y, a^*)$ is
defined as:
\begin{equation}
    R(y, a^*) = \lambda_{\mathrm{fmt}}\, R_{\mathrm{fmt}}(y)
              + \lambda_{\mathrm{acc}}\, R_{\mathrm{acc}}(y, a^*)
    \label{eq:reward}
\end{equation}
\noindent where $\lambda_{\mathrm{fmt}} + \lambda_{\mathrm{acc}} = 1$. Both $R_{\mathrm{fmt}}$ and $R_{\mathrm{acc}}$ are binary, so the reward takes one of four values, $R(y, a^*) \in \{0,\ \lambda_{\mathrm{fmt}},\ \lambda_{\mathrm{acc}},\ 1\}$. We set $(\lambda_{\mathrm{fmt}}, \lambda_{\mathrm{acc}}) = (0.3, 0.7)$ for the 3B model and $(0.2, 0.8)$ for the 7B model. 

\paragraph{Format Reward.}
The format reward $R_{\mathrm{fmt}}(y) \in \{0, 1\}$ verifies that the model has produced all four required structural tags in the rollout:
\begin{equation}
    R_{\mathrm{fmt}}(y) = \prod_{k=1}^{4} \mathbb{1}\bigl[\tau_k \in y\bigr]
    \label{eq:format}
\end{equation}
\noindent where $\{\tau_1, \tau_2, \tau_3, \tau_4\} = \{$\texttt{</extract\_entities>}, \texttt{</filtered\_groups>}, \texttt{</associative\_reasoning>}, \texttt{</answer>}$\}$ are the four closing tags corresponding to entity extraction, group filtering, associative reasoning, and the final answer, respectively. The reward is $1$ only when all four tags are present, ensuring the model completes the full three-turn pipeline.

\paragraph{Accuracy Reward.}
The accuracy reward $R_{\mathrm{acc}}(y, a^*) \in \{0, 1\}$ performs an exact match between the model's predicted answer and the ground truth:
\begin{equation}
    R_{\mathrm{acc}}(y, a^*) = \mathbb{1}\bigl[\mathrm{Extract}(y) = a^*\bigr]
    \label{eq:accuracy}
\end{equation}
\noindent where $\mathrm{Extract}(y)$ retrieves the text content between \texttt{<answer>} and \texttt{</answer>} tags in $y$, normalized to lowercase with whitespace stripped. If no \texttt{<answer>} tag is present, $\mathrm{Extract}(y) = \varnothing$ and the match fails.

\paragraph{Reward Design.}
The two components are combined additively so that a rollout completing the
full three-turn pipeline receives partial credit ($\lambda_{\mathrm{fmt}}$)
even when its final answer is wrong. This keeps a dense and easily satisfied
signal toward completing the pipeline throughout training: under an
outcome-only reward, a well-completed trajectory that reaches a wrong answer is
indistinguishable from one that breaks off midway, and the policy receives no
gradient toward following the multi-turn agentic structure. We place more weight on format for
the 3B model ($\lambda_{\mathrm{fmt}} = 0.3$ versus $0.2$ for 7B), as the
smaller base model has weaker instruction-following ability and needs a stronger
anchor on structural compliance.

\paragraph{Training Procedure.}
We train the policy with GRPO (Equation~\ref{eq:grpo-objective}) directly from the base instruct model, without any supervised fine-tuning stage. The additive reward removes the need for SFT initialization since the format term makes structural compliance rewarding from the very first rollouts, so that the instruction-tuned base model could acquire the multi-turn output format through RL alone, while the accuracy term drives answer quality in parallel. This purely RL-based training pipeline simplifies the overall training approach and demonstrates that the multi-turn agentic behavior can emerge entirely from reward-driven exploration.

\section{Experiments}
\label{sec::experiments}

The experiments in this section are designed to test the following 
hypotheses: (1) Knowledge extrapolation is a learnable behavior that 
is robust across model sizes, enabling small models to match or surpass 
large-model agentic frameworks (Section~\ref{sec::main_exp}). 
(2) The improvement stems from reasoning over structured knowledge 
rather than from fine-tuning alone (Section~\ref{sec::main_exp}). 
(3) SAKE generalizes across domains and KG scales, with the largest 
gains where knowledge is most incomplete (Section~\ref{sec::main_exp}). 
(4) SAKE achieves competitive accuracy while reducing token usage by 
over 90\% compared to inference-time agentic frameworks 
(Section~\ref{sec::efficiency_exp}). (5) Each component of the 
pipeline---key entity extraction, entity group construction, group filtering, agentic retrieval, and knowledge 
extrapolation---contributes meaningfully to the final performance 
(Section~\ref{sec::ablation_exp}).
\vspace{-0.2cm}
\subsection{Performance on Knowledge-Intensive Benchmarks}
\begin{table}[t]
\vspace{-0.5cm}
\centering
\captionsetup{font=footnotesize, labelfont=bf}
\caption{QA accuracy (\%) across two KGs of vastly different scales.
Agentic KG frameworks (GraphRAG, ToG, GIVE) are reported both on
GPT-3.5-Turbo, following the settings in which they were originally
published, and on the same Qwen2.5 backbones used by SAKE. Best results for
each model size are in \textbf{bold}. \textcolor{darkgreen}{$\Delta$}:
difference between SAKE and the strongest baseline \textit{within
the same model block}. }
\label{tab:main}
\renewcommand{\arraystretch}{1.12}
\footnotesize
\setlength{\tabcolsep}{2pt}
\begin{tabular*}{\columnwidth}{@{\extracolsep{\fill}} ll cccc c @{}}
\toprule
& & \multicolumn{4}{c}{\textbf{UMLS} {\scriptsize(135, 5.9K)}} & \textbf{CNet} {\scriptsize(844K, 2.1M)} \\
\cmidrule(lr){3-6} \cmidrule(lr){7-7}
\textbf{Model} & \textbf{Method} & \textbf{PubMedQA} & \textbf{BioASQ} & \textbf{ProcBank} & \textbf{Wt.Avg} & \textbf{CSQA} \\
\midrule
\multirow{3}{*}{\scriptsize\shortstack{GPT-3.5\\Turbo}}
  & GraphRAG  & 23.4 & 10.3 & 71.3 & 27.0 & ---  \\
  & ToG       & 17.6 & 18.0 & 66.8 & 26.6 & 69.8 \\
  & GIVE      & 53.6 & 88.2 & 73.4 & 70.1 & 74.7 \\
\midrule
\multirow{11}{*}{\scriptsize\shortstack{Qwen2.5\\3B-Instruct}}
  & I/O Prompt  & 11.0 & 28.6 & 67.5 & 27.7 & 64.2 \\
  & CoT         & 12.0 & 25.0 & 65.0 & 26.3 & 65.0 \\
  & RAG         & 22.0 & 20.2 & 52.5 & 26.8 & 52.0 \\
  & GraphRAG    & 17.0 &  9.5 & 65.0 & 22.8 & 63.8 \\
  & ToG         & 11.0 & 11.9 & 62.5 & 20.5 & 64.6 \\
  & GIVE        & 15.0 &  8.3 & 60.0 & 20.5 &  9.8 \\
  & SFT         & 33.0 & 76.2 & 55.0 & 53.1 & 67.5 \\
  & GRPO        & 19.0 & 69.1 & 67.5 & 46.5 & 66.7 \\
  & Search-R1   & 16.0 & 39.3 & 42.5 & 29.5 & 66.7 \\
\cmidrule(l){2-7}
  & \textbf{SAKE} & \textbf{51.0} & \textbf{82.1} & \textbf{70.0} & \textbf{66.1} & \textbf{73.2} \\
  & \multicolumn{1}{l}{\textcolor{darkgreen}{$\Delta$}} & \cellcolor{green!8}\textcolor{darkgreen}{+18.0} & \cellcolor{green!8}\textcolor{darkgreen}{+5.9} & \cellcolor{green!8}\textcolor{darkgreen}{+2.5} & \cellcolor{green!8}\textcolor{darkgreen}{+13.0} & \cellcolor{green!8}\textcolor{darkgreen}{+5.7} \\
\midrule
\multirow{11}{*}{\scriptsize\shortstack{Qwen2.5\\7B-Instruct}}
  & I/O Prompt  & 23.0 & 78.5 & 70.0 & 52.2 & 76.4 \\
  & CoT         & 29.0 & 77.4 & 75.0 & 55.4 & 63.4 \\
  & RAG         & 15.0 & 46.4 & 70.0 & 36.6 & 77.6 \\
  & GraphRAG    & 11.0 & 19.0 & \textbf{82.5} & 26.8 & 78.9 \\
  & ToG         & 10.0 & 14.3 & \textbf{82.5} & 24.6 & 77.2 \\
  & GIVE        & 14.0 & 44.0 & 47.5 & 31.2 & 14.6 \\
  & SFT         & 42.0 & 82.1 & 75.0 & 62.9 & 79.3 \\
  & GRPO        & 42.0 & 66.7 & 72.5 & 56.7 & 79.7 \\
  & Search-R1   & 24.0 & 78.6 & 75.0 & 53.6 & 78.5 \\
\cmidrule(l){2-7}
  & \textbf{SAKE} & \textbf{57.0} & \textbf{95.2} & 80.0 & \textbf{75.4} & \textbf{81.3} \\
  & \multicolumn{1}{l}{\textcolor{darkgreen}{$\Delta$}} & \cellcolor{green!8}\textcolor{darkgreen}{+15.0} & \cellcolor{green!8}\textcolor{darkgreen}{+13.1} & \cellcolor{red!8}\textcolor{red!70!black}{$-2.5$} & \cellcolor{green!8}\textcolor{darkgreen}{+12.5} & \cellcolor{green!8}\textcolor{darkgreen}{+1.6} \\
\bottomrule
\end{tabular*}
\par\vspace{2pt}
{\scriptsize CSQA = CommonsenseQA. CNet = ConceptNet (844K nodes, 2.1M edges).}
\vspace{-0.5cm}
\end{table}
\label{sec::main_exp}

To enable direct comparison with prior agentic KG reasoning methods, 
we evaluate on the same biomedical benchmarks and UMLS KG used in 
GIVE \citep{he2024give}: PubMedQA \citep{pubmedqa}, BioASQ 
\citep{bioasq}, and ProcessBank \citep{processbank}, with a 135-node 
UMLS subgraph \citep{umls_sparse} containing 5{,}877 triplets. We 
additionally evaluate on CommonsenseQA \citep{commonsenseqa} with 
ConceptNet \citep{conceptnet} to test generalization beyond the 
small-KG regime. For all datasets, we randomly select 80\% of the 
questions for training and evaluate on the remaining 20\%. 
Table~\ref{tab:main} presents the main results across these 
knowledge-intensive QA benchmarks using two KGs of vastly different 
scales. We organize our analysis around three key findings:

\textbf{Knowledge extrapolation is learnable and robust across 
model sizes.}
SAKE achieves the best weighted average on both Qwen2.5-3B and 7B, 
raising it
from 27.7\% to 66.1\% (+38.4\%) for 3B and from 52.2\% to 75.4\% 
(+23.2\%) for 7B. This consistency is notable because other fine-tuning 
baselines exhibit unstable gains across scales: SFT improves PubMedQA 
from 11.0\% to 33.0\% on 3B but only from 23.0\% to 42.0\% on 7B, 
while GRPO actually \textit{decreases} BioASQ accuracy on 7B (from 
78.5\% to 66.7\%). SAKE, by contrast, improves every benchmark on 
both model sizes without exception. More remarkably, these small 
models with SAKE surpass GPT-3.5-Turbo with GIVE, which is a state-of-the-art inference-time agentic framework that relies on multi-step prompting and iterative LLM calls. The 7B model achieves 75.4\% weighted average versus GIVE's 70.1\%, and even the 3B model (66.1\%) 
approaches GIVE's performance despite using a model with far fewer parameters. These results demonstrate that associative 
reasoning over structured knowledge, with linking entity groups and 
constructing new triplets by analogy, can be internalized by small 
models through RL, without complex multi-step instructions at 
inference time.
 
\textbf{The improvement stems from reasoning over structured knowledge, not from fine-tuning alone.}
To isolate the contribution of structured KG interaction, we compare SAKE against two fine-tuning baselines that use the same training data and optimization but lack access to KG tools. Vanilla GRPO trains with the same RL algorithm and reward signal but without KG retrieval: SAKE outperforms it by +19.6\% and +18.7\% weighted average accuracy on 3B and 7B, respectively. SFT, which directly fine-tunes on question-answer pairs, also falls short of SAKE by +13.0\% (3B) and +12.5\% (7B). These large, consistent gaps confirm that SAKE's gains are not an artifact of additional training compute or exposure to the training data, but arise specifically from learning to retrieve and reason over structured knowledge. This contrast highlights the advantage of structured KG retrieval: compact, relational, and directly amenable to associative reasoning, over unstructured text retrieval in knowledge-intensive scientific tasks.

\textbf{SAKE generalizes across domains and KG scales, with 
the largest gains where knowledge graph is most incomplete.}
To evaluate whether SAKE's learned extrapolation strategy transfers 
beyond the setting it was designed for, we apply it to CommonsenseQA 
using ConceptNet\citep{conceptnet}, a KG that is over 6{,}000$\times$ larger than UMLS 
and covers a fundamentally different domain (commonsense vs.\ 
biomedical). SAKE achieves the best accuracy on both model sizes 
(73.2\% on 3B, 81.3\% on 7B), confirming that the agentic pipeline 
generalizes without architectural or prompt modifications. However, 
the margins are notably smaller than on UMLS: +5.7\% (3B) and +1.6\% (7B) on CommonsenseQA versus +13.0\% (3B) and +12.5\% (7B) weighted average on UMLS. This pattern is consistent with SAKE's design: its distinctive advantage of extrapolating missing knowledge by analogy is the most impactful when the KG is small and incomplete, leaving less room for improvement when retrieval alone already provides adequate information coverage.
\vspace{-0.2cm}
\subsection{Efficiency Analysis}
\label{sec::efficiency_exp}

\begin{wraptable}{r}{0.52\columnwidth}
    \vspace{-12pt}
    \centering
    \captionsetup{font=footnotesize, labelfont=bf}
    \caption{Avg.\ token consumption per question. Reduction is vs.\ ToG / vs.\ GIVE.}
    \label{tab:token}
    \renewcommand{\arraystretch}{1.1}
    \scriptsize
    \setlength{\tabcolsep}{3pt}
    \begin{tabular}{@{}l ccc@{}}
    \toprule
    \textbf{Method} & \textbf{PubMedQA} & \textbf{BioASQ} & \textbf{ProcBank} \\
    \midrule
    ToG   & 12{,}701 & 7{,}010 & 11{,}996 \\
    GIVE  & 14{,}519 & 7{,}970 & 19{,}461 \\
    \midrule
    \textbf{SAKE} & \textbf{800} & \textbf{639} & \textbf{1{,}156} \\
    \rowcolor{green!8}
    Reduct. & \textcolor{darkgreen}{\scriptsize 93.7/94.5} & \textcolor{darkgreen}{\scriptsize 90.9/92.0} & \textcolor{darkgreen}{\scriptsize 90.4/94.1} \\
    \bottomrule
    \end{tabular}
    \vspace{-10pt}
\end{wraptable}

A key advantage of SAKE over inference-time agentic frameworks is computational efficiency. Agentic methods such as ToG and GIVE involve multiple sequential LLM calls, where the output of each call is parsed as the input of the next. We measure \textit{total token usage} as the sum of input tokens across all LLM calls required to answer a single question---this captures the cumulative cost of iterative refinement. As shown in Table~\ref{tab:token}, ToG and GIVE consume 7{,}000--19{,}000 tokens per question due to repeated invocations for entity pruning, relation verification, and progressive answer generation. SAKE requires only a single inference pass: the query, tool-returned entity groups, and triplets are concatenated into one context, reducing total token usage by over 90\% across all datasets.
\subsection{Ablation Studies}
\label{sec::ablation_exp}
 
We conduct ablation experiments to isolate the contribution of SAKE's three pipeline components: group filtering, agentic retrieval, and knowledge extrapolation, by comparing the full pipeline against variants that remove or replace each one. We also study the effect of the number of additional entities per group ($p$), finding that $p=3$ yields optimal performance; this hyperparameter analysis is deferred to Appendix~\ref{appendix:hp}. Unless otherwise stated, all ablation experiments use the three biomedical benchmarks (UMLS KG) with Qwen2.5-7B-Instruct, as the scale of ConceptNet makes repeated ablation runs prohibitively expensive.

\subsubsection{Contribution of Pipeline Components}

\begin{figure}[t]
    \centering
    \includegraphics[width=\columnwidth]{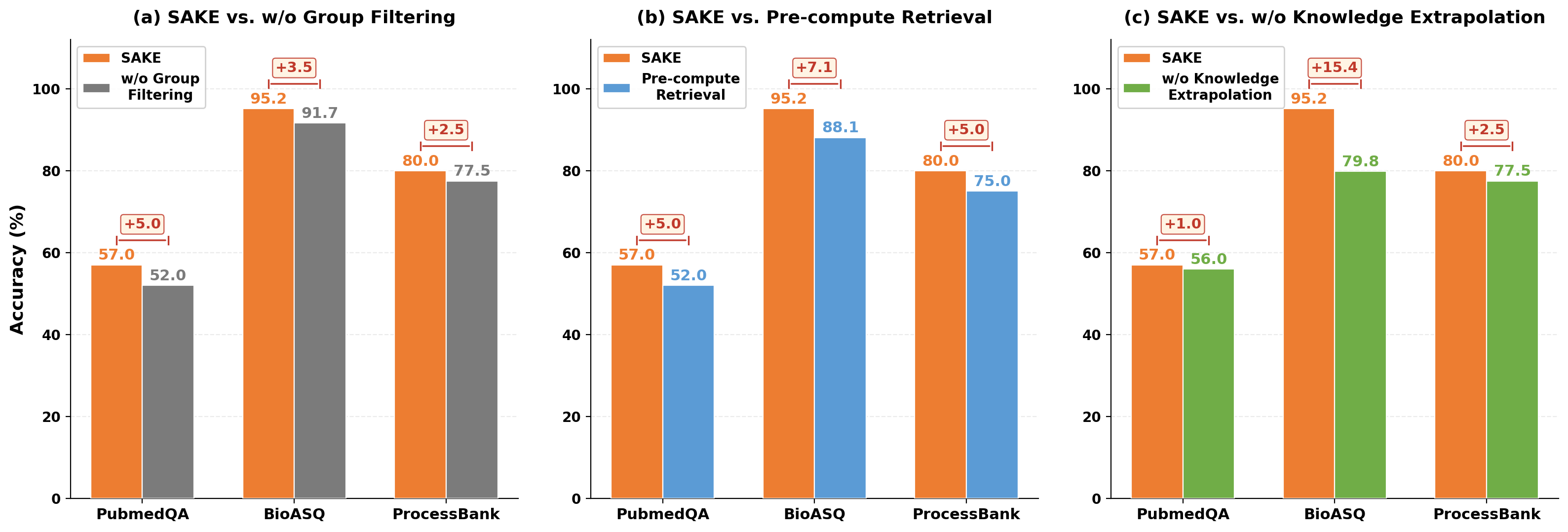}
    \captionsetup{font=footnotesize, labelfont=bf}
    \caption{Component ablation on Qwen2.5-7B (Biomedical QA). (a) Full pipeline vs.\ removing the group filtering turn. Red annotations show the accuracy gain from each component. (b) Agentic tool use vs.\ pre-computed retrieval. (c) Full pipeline vs.\ removing the knowledge extrapolation turn. }
    \label{fig::ablation_components}
    \vspace{-0.3cm}
\end{figure}

Figure~\ref{fig::ablation_components} isolates the contribution of three key design choices in SAKE by comparing the full pipeline against variants that remove or replace each component.

\textbf{Group filtering reduces noise from irrelevant entity groups.}
In panel (a), we remove the group filtering turn: all entity groups returned by Tool~1 are passed directly to Tool~2 without the model selecting which to retain. This variant decreases accuracy by 5.0\% on PubMedQA, 3.5\% on BioASQ, and 2.5\% on ProcessBank. The filtering step matters because entity extraction may produce concepts whose KG neighborhoods are tangential to the question. Without filtering, Tool~2 retrieves cross-group triplets involving these irrelevant groups, introducing noise into the reasoning context. By learning to discard low-quality groups before triplet retrieval, the model ensures that the knowledge it reasons over is focused and relevant, improving the precision of the subsequent extrapolation step.

\textbf{Agentic retrieval outperforms pre-computed retrieval.}
In panel (b), we compare SAKE's learned agentic retrieval against a variant where entity groups and KG triplets are pre-computed offline and appended to the prompt before training, as in prior work. SAKE outperforms the pre-computed variant on all three benchmarks: +5.0\% on PubMedQA, +7.1\% on BioASQ, and +5.0\% on ProcessBank. The advantage of agentic retrieval is that the model has the freedom to decide \textit{what} to extract and retrieve conditioned on the specific query, rather than receiving a fixed set of knowledge determined by an external pipeline. This allows the model to adapt its retrieval strategy to each question: extracting different entities, retaining different groups, and then tailor its reasoning accordingly. The end-to-end reward signal reinforces this adaptive behavior, producing retrieval decisions that are jointly optimized with the downstream reasoning.

\textbf{Knowledge extrapolation is essential for bridging incomplete knowledge.}
In panel (c), we remove the associative reasoning turn by eliminating the \texttt{<associative\_reasoning>} tag from the prompt: after retrieving KG triplets, the model directly generates an answer without an explicit extrapolation step. This ablation degrades performance across all benchmarks, with the largest drop on BioASQ ($-15.4$\%), followed by ProcessBank ($-2.5$\%) and PubMedQA ($-1.0$\%). The BioASQ result is particularly revealing: this dataset contains many questions whose answers are not directly present in the sparse UMLS KG, making the constructing new triplets by substituting semantically similar concepts critical for bridging the gap. Without it, the model must reason directly from retrieved triplets that may not connect to the query, leading to substantially lower accuracy.

\vspace{-0.2cm}
\section{Conclusion}
\label{sec::conclusion}

We proposed SAKE, an agentic framework that trains LLMs to perform knowledge extrapolation over structured knowledge graphs through tool-augmented reinforcement learning. SAKE defines a three-turn pipeline, including entity extraction, group filtering, and associative reasoning. This rollout generation process is interleaved with two deterministic KG tool calls, all optimized end-to-end with GRPO using a single outcome-based reward. With SAKE, a Qwen2.5-7B model surpasses GPT-3.5-Turbo with state-of-the-art agentic KG traverse frameworks on both biomedical and commonsense benchmarks, while reducing token usage by over 90\%. These results demonstrate that associative reasoning over limited external knowledge is a learnable behavior that small, open-weight models can acquire through reinforcement learning, and we hope SAKE encourages further exploration of tool-augmented RL for deploying knowledge-grounded reasoning in specialized domains where comprehensive knowledge bases are impractical to maintain.
\bibliography{custom}

@article{BEATY2023671,
title = {Associative thinking at the core of creativity},
journal = {Trends in Cognitive Sciences},
volume = {27},
number = {7},
pages = {671-683},
year = {2023},
issn = {1364-6613},
doi = {https://doi.org/10.1016/j.tics.2023.04.004},
url = {https://www.sciencedirect.com/science/article/pii/S1364661323000943},
author = {Roger E. Beaty and Yoed N. Kenett}
}

@article{KAUFMAN2009374,
title = {Associative learning predicts intelligence above and beyond working memory and processing speed},
journal = {Intelligence},
volume = {37},
number = {4},
pages = {374-382},
year = {2009},
issn = {0160-2896},
doi = {https://doi.org/10.1016/j.intell.2009.03.004},
url = {https://www.sciencedirect.com/science/article/pii/S0160289609000300},
author = {Scott Barry Kaufman and Colin G. DeYoung and Jeremy R. Gray and Jamie Brown and Nicholas Mackintosh}
}

@article{he2024give,
  title={GIVE: Structured Reasoning with Knowledge Graph Inspired Veracity Extrapolation},
  author={He, Jiashu and Ma, Mingyu Derek and Fan, Jinxuan and Roth, Dan and Wang, Wei and Ribeiro, Alejandro},
  journal={arXiv preprint arXiv:2410.08475},
  year={2024}
}

@misc{shao2024deepseekmathpushinglimitsmathematical,
      title={DeepSeekMath: Pushing the Limits of Mathematical Reasoning in Open Language Models}, 
      author={Zhihong Shao and Peiyi Wang and Qihao Zhu and Runxin Xu and Junxiao Song and Xiao Bi and Haowei Zhang and Mingchuan Zhang and Y. K. Li and Y. Wu and Daya Guo},
      year={2024},
      eprint={2402.03300},
      archivePrefix={arXiv},
      primaryClass={cs.CL},
      url={https://arxiv.org/abs/2402.03300}, 
}

@article{cascella2023evaluating,
  title={Evaluating the feasibility of ChatGPT in healthcare: an analysis of multiple clinical and research scenarios},
  author={Cascella, Marco and Montomoli, Jonathan and Bellini, Valentina and Bignami, Elena},
  journal={Journal of medical systems},
  volume={47},
  number={1},
  pages={33},
  year={2023},
  publisher={Springer}
}

@article{tian2024opportunities,
  title={Opportunities and challenges for ChatGPT and large language models in biomedicine and health},
  author={Tian, Shubo and Jin, Qiao and Yeganova, Lana and Lai, Po-Ting and Zhu, Qingqing and Chen, Xiuying and Yang, Yifan and Chen, Qingyu and Kim, Won and Comeau, Donald C and others},
  journal={Briefings in Bioinformatics},
  volume={25},
  number={1},
  pages={bbad493},
  year={2024},
  publisher={Oxford University Press}
}

@misc{sun2024thinkongraphdeepresponsiblereasoning,
      title={Think-on-Graph: Deep and Responsible Reasoning of Large Language Model on Knowledge Graph}, 
      author={Jiashuo Sun and Chengjin Xu and Lumingyuan Tang and Saizhuo Wang and Chen Lin and Yeyun Gong and Lionel M. Ni and Heung-Yeung Shum and Jian Guo},
      year={2024},
      eprint={2307.07697},
      archivePrefix={arXiv},
      primaryClass={cs.CL},
      url={https://arxiv.org/abs/2307.07697}, 
}

@misc{edge2025localglobalgraphrag,
      title={From Local to Global: A Graph RAG Approach to Query-Focused Summarization}, 
      author={Darren Edge and Ha Trinh and Newman Cheng and Joshua Bradley and Alex Chao and Apurva Mody and Steven Truitt and Dasha Metropolitansky and Robert Osazuwa Ness and Jonathan Larson},
      year={2025},
      eprint={2404.16130},
      archivePrefix={arXiv},
      primaryClass={cs.CL},
      url={https://arxiv.org/abs/2404.16130}, 
}

@misc{searchO1,
      title={Search-o1: Agentic Search-Enhanced Large Reasoning Models}, 
      author={Xiaoxi Li and Guanting Dong and Jiajie Jin and Yuyao Zhang and Yujia Zhou and Yutao Zhu and Peitian Zhang and Zhicheng Dou},
      year={2025},
      eprint={2501.05366},
      archivePrefix={arXiv},
      primaryClass={cs.AI},
      url={https://arxiv.org/abs/2501.05366}, 
}

@article{umls_sparse,
author = {Li, Da and Zhu, Boqing and Yang, Sen and Xu, Kele and Yi, Ming and He, Yukai and Wang, Huaimin},
title = {Multi-task Pre-training Language Model for Semantic Network Completion},
year = {2023},
issue_date = {November 2023},
publisher = {Association for Computing Machinery},
address = {New York, NY, USA},
volume = {22},
number = {11},
issn = {2375-4699},
url = {https://doi.org/10.1145/3627704},
doi = {10.1145/3627704},
journal = {ACM Trans. Asian Low-Resour. Lang. Inf. Process.},
month = {nov},
articleno = {250},
numpages = {20}
}

@inproceedings{pubmedqa,
  title={PubMedQA: A Dataset for Biomedical Research Question Answering},
  author={Jin, Qiao and Dhingra, Bhuwan and Liu, Zhengping and Cohen, William and Lu, Xinghua},
  booktitle={Proceedings of the 2019 Conference on Empirical Methods in Natural Language Processing and the 9th International Joint Conference on Natural Language Processing (EMNLP-IJCNLP)},
  pages={2567--2577},
  year={2019}
}

@article{bioasq,
	title = {BioASQ-QA: A manually curated corpus for Biomedical Question Answering},
	journal = {Scientific Data},
	volume = {10},
	year = {2023},
	pages = {170},
	url = {https://doi.org/10.1038/s41597-023-02068-4},
	author = {Krithara, Anastasia and Nentidis, Anastasios and Bougiatiotis, Konstantinos and Paliouras, Georgios}
}

@inproceedings{processbank,
    title = "Modeling Biological Processes for Reading Comprehension",
    author = "Berant, Jonathan  and
      Srikumar, Vivek  and
      Chen, Pei-Chun  and
      Vander Linden, Abby  and
      Harding, Brittany  and
      Huang, Brad  and
      Clark, Peter  and
      Manning, Christopher D.",
    editor = "Moschitti, Alessandro  and
      Pang, Bo  and
      Daelemans, Walter",
    booktitle = "Proceedings of the 2014 Conference on Empirical Methods in Natural Language Processing ({EMNLP})",
    month = oct,
    year = "2014",
    address = "Doha, Qatar",
    publisher = "Association for Computational Linguistics",
    url = "https://aclanthology.org/D14-1159",
    doi = "10.3115/v1/D14-1159",
    pages = "1499--1510",
}

@misc{searchR1,
      title={Search-R1: Training LLMs to Reason and Leverage Search Engines with Reinforcement Learning}, 
      author={Bowen Jin and Hansi Zeng and Zhenrui Yue and Jinsung Yoon and Sercan Arik and Dong Wang and Hamed Zamani and Jiawei Han},
      year={2025},
      eprint={2503.09516},
      archivePrefix={arXiv},
      primaryClass={cs.CL},
      url={https://arxiv.org/abs/2503.09516}, 
}

@inproceedings{verl, series={EuroSys ’25},
   title={HybridFlow: A Flexible and Efficient RLHF Framework},
   url={http://dx.doi.org/10.1145/3689031.3696075},
   DOI={10.1145/3689031.3696075},
   booktitle={Proceedings of the Twentieth European Conference on Computer Systems},
   publisher={ACM},
   author={Sheng, Guangming and Zhang, Chi and Ye, Zilingfeng and Wu, Xibin and Zhang, Wang and Zhang, Ru and Peng, Yanghua and Lin, Haibin and Wu, Chuan},
   year={2025},
   month=mar, pages={1279–1297},
   collection={EuroSys ’25} }

@misc{conceptnet,
      title={ConceptNet 5.5: An Open Multilingual Graph of General Knowledge}, 
      author={Robyn Speer and Joshua Chin and Catherine Havasi},
      year={2018},
      eprint={1612.03975},
      archivePrefix={arXiv},
      primaryClass={cs.CL},
      url={https://arxiv.org/abs/1612.03975}, 
}

@misc{commonsenseqa,
      title={CommonsenseQA: A Question Answering Challenge Targeting Commonsense Knowledge}, 
      author={Alon Talmor and Jonathan Herzig and Nicholas Lourie and Jonathan Berant},
      year={2019},
      eprint={1811.00937},
      archivePrefix={arXiv},
      primaryClass={cs.CL},
      url={https://arxiv.org/abs/1811.00937}, 
}

@misc{luo2025graphr1,
      title={Graph-R1: Towards Agentic GraphRAG Framework via End-to-end Reinforcement Learning}, 
      author={Haoran Luo and Haihong E and Guanting Chen and Qika Lin and Yikai Guo and Fangzhi Xu and Zemin Kuang and Meina Song and Xiaobao Wu and Yifan Zhu and Luu Anh Tuan},
      year={2025},
      eprint={2507.21892},
      archivePrefix={arXiv},
      primaryClass={cs.CL},
      url={https://arxiv.org/abs/2507.21892}, 
}

@misc{tsang2025autographr,
      title={AutoGraph-R1: End-to-End Reinforcement Learning for Knowledge Graph Construction}, 
      author={Hong Ting Tsang and Jiaxin Bai and Haoyu Huang and Qiao Xiao and Tianshi Zheng and Baixuan Xu and Shujie Liu and Yangqiu Song},
      year={2025},
      eprint={2510.15339},
      archivePrefix={arXiv},
      primaryClass={cs.CL},
      url={https://arxiv.org/abs/2510.15339}, 
}

@misc{knowledgeabler1,
      title={Resisting Contextual Interference in RAG via Parametric-Knowledge Reinforcement}, 
      author={Chenyu Lin and Yilin Wen and Du Su and Hexiang Tan and Fei Sun and Muhan Chen and Chenfu Bao and Zhonghou Lyu},
      year={2026},
      eprint={2506.05154},
      archivePrefix={arXiv},
      primaryClass={cs.CL},
      url={https://arxiv.org/abs/2506.05154}, 
}

@inproceedings{xiong2025deliberate,
  title={Deliberate reasoning in language models as structure-aware planning with an accurate world model},
  author={Xiong, Siheng and Payani, Ali and Yang, Yuan and Fekri, Faramarz},
  booktitle={Proceedings of the 63rd Annual Meeting of the Association for Computational Linguistics (Volume 1: Long Papers)},
  pages={31900--31931},
  year={2025}
}

@article{xiong2025enhancing,
  title={Enhancing Long Chain-of-Thought Reasoning through Multi-Path Plan Aggregation},
  author={Xiong, Siheng and Payani, Ali and Fekri, Faramarz},
  journal={arXiv preprint arXiv:2510.11620},
  year={2025}
}
\bibliographystyle{colm2026_conference}

\clearpage
\appendix
\label{sec:appendix}
\section{Dataset Statistics}
\begin{table}[H]
\centering
\captionsetup{font=footnotesize, labelfont=bf}
\caption{Summary of dataset statistics. $|\mathcal{V}|$ and $|\mathcal{E}|$ are the
number of nodes and edges in the associated knowledge graph. \#Test is the number of
held-out questions on which all methods are evaluated; the weighted average reported
in Table~\ref{tab:main} pools the three UMLS datasets over their combined 224 questions.}
\label{tab:dataset}
\renewcommand{\arraystretch}{1.15}
\footnotesize
\setlength{\tabcolsep}{4pt}
\begin{tabular}{@{} l r r l l r @{}}
\toprule
\textbf{KG} & $|\mathcal{V}|$ & $|\mathcal{E}|$ & \textbf{Dataset} & \textbf{QA Type} & \textbf{\#Test} \\
\midrule
\multirow{4}{*}{UMLS \citep{umls_sparse}}
  & \multirow{4}{*}{135}
  & \multirow{4}{*}{5,877}
  & PubMedQA \citep{pubmedqa}       & Yes--No       & 100 \\
  & & & BioASQ \citep{bioasq}            & Yes--No       &  84 \\
  & & & ProcessBank \citep{processbank}  & Multi-Choice  &  40 \\
\cmidrule(l){4-6}
  & & & \textit{UMLS total}              &               & \textbf{224} \\
\midrule
CNet \citep{conceptnet}
  & 844,158
  & 2,085,099
  & CSQA \citep{commonsenseqa} & Multi-Choice & 246 \\
\bottomrule
\end{tabular}
\end{table}
Table \ref{tab:dataset} summarizes the statistics of all datasets used in Section \ref{sec::main_exp}.

\label{appendix:dataset}

\section{Computing infrastructures}
We ran experiments on a workstation with 2 Nvidia A6000 GPUs, and an AMD Ryzen Threadripper 3960X CPU (24C/48T, x86\_64, 2.2–3.8 GHz, single NUMA node) with 768 KiB L1d + 768 KiB L1i, 12 MiB L2, and 128 MiB L3 cache; AMD-V enabled.

\section{Number of Additional Entities per Group}
\label{appendix:hp}
\begin{wrapfigure}{r}{0.5\columnwidth}
    \vspace{-12pt}
    \centering
    \includegraphics[width=0.5\columnwidth]{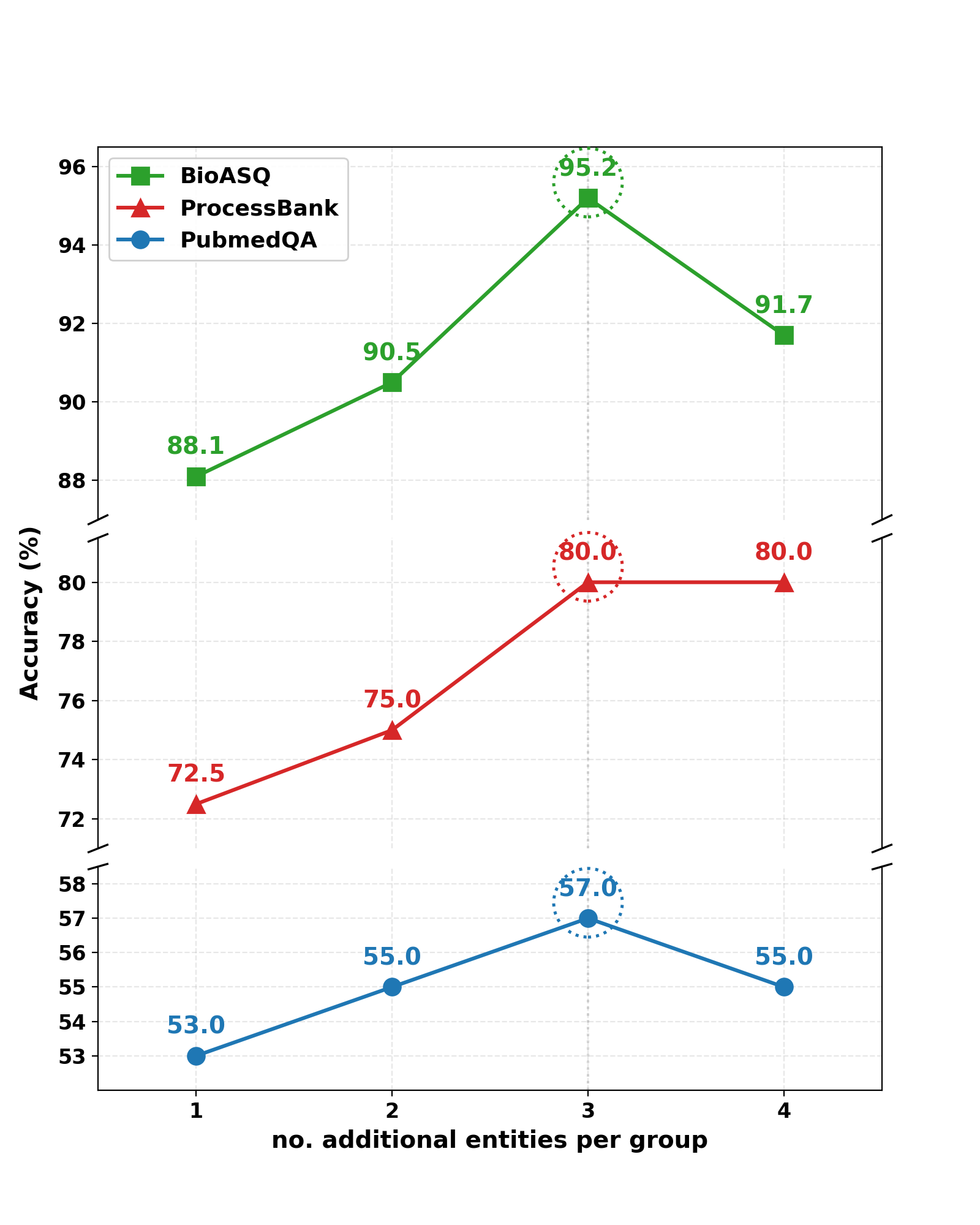}
    \captionsetup{font=footnotesize, labelfont=bf}
    \caption{QA accuracy of SAKE (Qwen2.5-7B) with varying numbers of additional entities per group ($p$ in Eq.~\ref{eq:group-construction}). Dotted circles mark the best setting ($p=3$).}
    \label{fig::ablation_entities}
    \vspace{-10pt}
\end{wrapfigure}

The number of additional entities per group ($p$ in Eq.~\ref{eq:group-construction}) controls how broadly each entity group extends beyond the original query concept. A small $p$ restricts each group to the most semantically similar KG concepts, yielding precise but narrow coverage. A large $p$ includes more distant concepts, broadening coverage at the risk of introducing noise.

Figure~\ref{fig::ablation_entities} shows the effect of varying $p$ from 1 to 4 on Qwen2.5-7B across the three UMLS benchmarks. Performance improves steadily as $p$ increases from 1 to 3, with all three datasets reaching their peak at $p=3$: BioASQ achieves 95.2\%, ProcessBank 80.0\%, and PubMedQA 57.0\%. At $p=4$, performance declines on BioASQ ($-3.5$) and PubMedQA ($-2.0$), while ProcessBank plateaus. This pattern reflects a trade-off inherent to knowledge extrapolation: too few additional entities leave the model with insufficient KG coverage to bridge the gap between query concepts and available triplets, while too many introduce semantically distant concepts that dilute the entity groups and lead to retrieval of irrelevant triplets, degrading the quality of associative reasoning.

\clearpage
\section{Rollout Algorithm}
\label{appendix:algo}
\definecolor{stoptok}{RGB}{192,0,0}       
\definecolor{tooltok}{RGB}{0,128,0}       
\definecolor{modeltok}{RGB}{128,0,128}    
\definecolor{sectioncolor}{RGB}{0,80,160} 

\begin{figure}[H]
\captionsetup{name=Algorithm}
\begin{minipage}{\columnwidth}
\hrule height 0.5pt \vspace{4pt}
\textbf{Algorithm 1} Multi-Turn KG-Guided Reasoning Rollout \\[-2pt]
\hrule height 0.5pt \vspace{2pt}
\begin{algorithmic}[1]
    \Require Input query $x$, policy model $\pi_\theta$, KG engine $\mathcal{K}$, similarity function $\mathrm{Sim}(\cdot)$, number of similar concepts $p$

    \State Initialize empty rollout sequence $y$; \; token mask $m \leftarrow []$

    \Statex \hspace{-\algorithmicindent} \textcolor{sectioncolor}{\textbf{$\triangleright$ Turn 1: Entity Extraction}}
    \Repeat
        \State Generate token $y_t \sim \pi_\theta(\cdot \mid x, y)$; \; $y \leftarrow y + y_t$; \; $m \leftarrow m + [1]$
    \Until{$y_t = $ \textcolor{stoptok}{\texttt{</extract\_entities>}}}
    \State $E = \{e_1, \ldots, e_n\} \leftarrow \mathrm{Parse}(y,$ \textcolor{stoptok}{\texttt{<extract\_entities>}}$)$

    \Statex \hspace{-\algorithmicindent} \textcolor{sectioncolor}{\textbf{$\triangleright$ Tool Call 1: Entity Group Construction} \textit{(no gradient)}}
    \For{each $e_i \in E$}
        \State $G_i \leftarrow \{e_i\} \cup \mathrm{Top\text{-}}p\;\mathrm{Sim}(e_i, \mathcal{K}.\mathrm{concepts})$
    \EndFor
    \State $o_1 \leftarrow$ \textcolor{tooltok}{\texttt{<entity\_groups>}} $\{G_1, \ldots, G_n\}$ \textcolor{tooltok}{\texttt{</entity\_groups>}}
    \State $y \leftarrow y + o_1$; \; $m \leftarrow m + [\mathbf{0}]^{|o_1|}$ \Comment{Mask tool tokens}

    \Statex \hspace{-\algorithmicindent} \textcolor{sectioncolor}{\textbf{$\triangleright$ Turn 2: Group Filtering}}
    \Repeat
        \State Generate token $y_t \sim \pi_\theta(\cdot \mid x, y)$; \; $y \leftarrow y + y_t$; \; $m \leftarrow m + [1]$
    \Until{$y_t = $ \textcolor{stoptok}{\texttt{</filtered\_groups>}}}
    \State $S \subseteq \{1, \ldots, n\} \leftarrow \mathrm{Parse}(y,$ \textcolor{stoptok}{\texttt{<filtered\_groups>}}$)$ \Comment{Selected group IDs}

    \Statex \hspace{-\algorithmicindent} \textcolor{sectioncolor}{\textbf{$\triangleright$ Tool Call 2: KG Triplet Retrieval} \textit{(no gradient)}}
    \State $T \leftarrow \varnothing$
    \For{each pair $(i, j) \in \binom{S}{2}$}
        \State $T \leftarrow T \cup \{(h, r, t) \in \mathcal{K}.\mathrm{edges} \mid h \in G_i, t \in G_j\}$ \Comment{Forward}
        \State $T \leftarrow T \cup \{(h, r, t) \in \mathcal{K}.\mathrm{edges} \mid h \in G_j, t \in G_i\}$ \Comment{Reverse}
    \EndFor
    \State $o_2 \leftarrow$ \textcolor{tooltok}{\texttt{<kg\_triplets>}} $T$ \textcolor{tooltok}{\texttt{</kg\_triplets>}}
    \State $y \leftarrow y + o_2$; \; $m \leftarrow m + [\mathbf{0}]^{|o_2|}$ \Comment{Mask tool tokens}

    \Statex \hspace{-\algorithmicindent} \textcolor{sectioncolor}{\textbf{$\triangleright$ Turn 3: Knowledge Extrapolation \& Answer}}
    \Repeat \Comment{Model generates reasoning + answer until EOS}
        \State Generate token $y_t \sim \pi_\theta(\cdot \mid x, y)$; \; $y \leftarrow y + y_t$; \; $m \leftarrow m + [1]$
    \Until{$y_t = $ \texttt{<eos>}}

    \State \Return rollout $y$, mask $m$
    \end{algorithmic}
\vspace{2pt}\hrule height 0.5pt
\end{minipage}
\vspace{6pt}
\captionsetup{font=footnotesize}
\caption*{Multi-turn rollout procedure for SAKE. The policy $\pi_\theta$ 
generates tokens across three turns, interleaved with two deterministic 
tool calls (highlighted in green). A binary mask $m$ tracks which tokens 
are model-generated ($1$) versus tool-injected ($0$), ensuring that 
GRPO loss is computed only over the model's own outputs.}
\end{figure}

\clearpage
\section{SAKE System Prompt}
\label{appendix:prompt}
\begin{table}[H]
\centering
\begin{tcolorbox}[examplebox, title=SAKE System Prompt,breakable]
\footnotesize\ttfamily
You are a biomedical reasoning assistant. To answer questions, you follow a structured retrieval and reasoning process using a knowledge graph. Follow these steps exactly.

\medskip
\textrm{\textbf{--- STEP 1: Entity Extraction ---}}

Think about the question. Identify the key substantive biomedical concepts --- these are specific biological entities, diseases, molecules, or processes. Do NOT extract relational or intent words (for example, do not extract "treatment", "cause", "effect", "role", "relationship", "association", "mechanism", "involvement", "function").

Format your response as:

<think> [your reasoning about which concepts are substantive] </think>

<extract\_entities> concept\_1 | concept\_2 | ... </extract\_entities>

\medskip
\textrm{\textbf{--- STEP 2: Group Filtering ---}}

You will receive entity groups: for each concept you extracted, a set of semantically related terms from the knowledge graph.

Review the groups. Keep the groups that are relevant to answering the question. Reference groups by their number.

Format your response as:

<think> [your reasoning about which groups to keep and why] </think>

<filtered\_groups> 1 | 2 | ... </filtered\_groups>

\medskip
\textrm{\textbf{--- STEP 3: Associative Reasoning ---}}

You will receive knowledge graph triplets connecting terms across your selected groups.

Use these triplets to reason associatively:
\begin{itemize}[nosep,leftmargin=*]
\item Identify which triplet relationships are relevant to the question
\item If a direct answer is not present, construct new triplets by substituting semantically similar terms from the entity groups into existing triplet patterns. For example: if (hormone, treats, mental\_disorder) is retrieved and melatonin is in the hormone group and insomnia is in the mental\_disorder group, construct (melatonin, treats, insomnia) by analogy.
\item Build a rationale connecting the retrieved knowledge to the question
\end{itemize}

Your answer must be a single word: yes, no, maybe, a, or b.

Format your response as:

<associative\_reasoning> your reasoning </associative\_reasoning>

<answer> your answer </answer>

\medskip
If no knowledge graph triplets are found, reason using the entity groups alone combined with your general knowledge.

\medskip
Question: \{question\}
\end{tcolorbox}
\captionsetup{font=footnotesize, labelfont=bf}
\caption{System prompt for SAKE, used identically during training and 
inference. The prompt defines the three-turn agentic pipeline and the 
expected output format for each turn, including the special tokens that 
trigger external tool calls. No task-specific examples or few-shot 
demonstrations are provided; the model's reasoning ability emerges 
entirely from RL training. \texttt{\{question\}} is replaced with the 
specific query at runtime.}
\label{tab:template}
\end{table}

\clearpage

\section{Training Analysis}
\label{appendix:training}

\begin{figure}[h]
    \centering
    \includegraphics[width=\columnwidth]{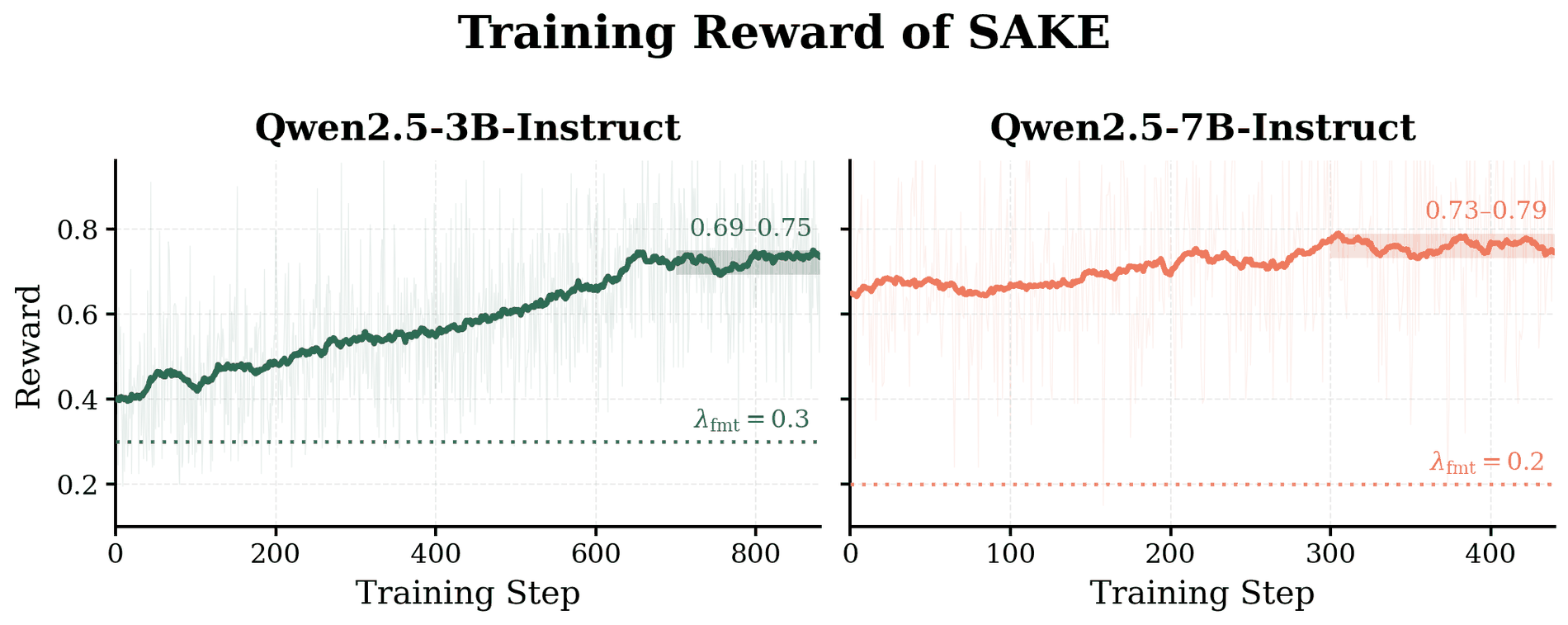}
    \captionsetup{font=footnotesize, labelfont=bf}
    \caption{Training reward for SAKE on Qwen2.5-3B-Instruct (left) and
    Qwen2.5-7B-Instruct (right), shown up to the checkpoint reported for
    each model. Light lines show per-step batch means; bold lines show a $50$-step
    centered moving average. Shaded bands mark the range over which the
    smoothed reward fluctuates after convergence.}
    \label{fig:traininglog}
\end{figure}

Figure~\ref{fig:traininglog} shows the training reward for SAKE on
Qwen2.5-3B-Instruct and Qwen2.5-7B-Instruct, up to the checkpoint
reported for each model. We log only the total reward, which combines
the format and accuracy components (Equation~\ref{eq:reward}), so the
curves describe overall training progress rather than the behavior of
either component individually.

The 3B model improves gradually, rising from approximately $0.40$ to a
plateau of $0.69$--$0.75$. The 7B model starts higher and improves
faster, converging to $0.73$--$0.79$. Both models begin well above
their respective $\lambda_{\mathrm{fmt}}$ levels, indicating that the
instruction-tuned base models already produce a substantial fraction of
well-formed rollouts before any policy updates.

Per-step batch means are noisy for both models, with individual steps
ranging from below $0.2$ to $1.0$. The curves in
Figure~\ref{fig:traininglog} are therefore smoothed with a $50$-step
centered moving average.

\section{Baseline Prompts and examples}\label{sec:prompt}
In this Section we provide detailed prompts for SAKE and each competing baselines during training and testing.
\subsection{I/O Prompt}

\begin{tcolorbox}[examplebox,title=I/O Prompt User Message]
\footnotesize\ttfamily
Answer the given question. Provide your answer inside <answer> and </answer>.
Question: \textbf{Q}. Answer this question with yes, no or maybe.
\end{tcolorbox}

\subsection{CoT Prompt}

\begin{tcolorbox}[examplebox,title=CoT Prompt User Message]
\footnotesize\ttfamily
Answer the given question. Provide your answer inside <answer> and </answer>. Question: \textbf{Q}. Answer this question with yes, no or maybe. Let's think step by step.            
\end{tcolorbox}

\subsection{RAG Prompt}

\begin{tcolorbox}[examplebox,title=RAG Prompt User Message]
\footnotesize\ttfamily
Answer the given question based on the retrieved knowledge triplets (entity, relation, entity) and your own knowledge. Provide your answer inside <answer> and </answer>. Question: \textbf{Q}. Answer this question with yes, no or maybe. Knowledge triplets: \textbf{Retrieved knowledge triplets} \end{tcolorbox}

\subsection{SFT Prompt}

\begin{tcolorbox}[examplebox,title=SFT  Prompt User Message]
\footnotesize\ttfamily
Answer the given question. Provide your answer inside <answer> and </answer>.
Question: \textbf{Q}. Answer this question with yes, no or maybe.

\end{tcolorbox}

\subsection{GRPO Prompt}

\begin{tcolorbox}[examplebox,title=GRPO  Prompt User Message]
\footnotesize\ttfamily
Answer the given question. You must conduct reasoning first inside <think> and </think>. After reasoning, provide your final answer inside <answer> and </answer>. Question: \textbf{Q}. Answer this question with yes, no or maybe.

\end{tcolorbox}

\end{document}